\RequirePackage{amsmath}
\documentclass[a4paper]{svproc}
\usepackage[utf8]{inputenc}
\usepackage[T1]{fontenc}

\usepackage{ntheorem}
\newtheorem{hyp}{Hypothesis}

\usepackage{conventions}
\usepackage{todonotes}
\usepackage{enumitem}
\usepackage{booktabs}
\usepackage{float}
\usepackage{wrapfig}
\usepackage{subfigure}
\usepackage{graphicx}
\graphicspath{ {./figures/} }

\usepackage{algorithm}
\usepackage{algpseudocode}

\usepackage{url}

\usepackage{import}

\usepackage{pgf}
\usepackage{pgfplots}
\pgfplotsset{compat=1.13}

\usepackage{xcolor}

\usepackage[normalem]{ulem}

\usepackage[style=ieee, backend=biber, natbib=true]{biblatex}
\renewcommand*{\bibfont}{\normalfont\scriptsize}
\newcommand{\customsubsubsection}[1]{\textbf{#1}}

\newcommand\blfootnote[1]{%
  \begingroup
  \renewcommand\thefootnote{}\footnote{#1}%
  \addtocounter{footnote}{-1}%
  \endgroup
}

\title{Inferring Occluded Geometry Improves Performance when Retrieving an Object from Dense Clutter}
\titlerunning{Inferring Occluded Geometry}
\author{Andrew Price\thanks{Equal contribution.} \and Linyi Jin$^\star$  \and Dmitry Berenson}
\authorrunning{Andrew Price, Linyi Jin, and Dmitry Berenson}
\institute{University of Michigan, Ann Arbor, MI 48109, USA\\ \email{\{pricear, jinlinyi\}@umich.edu, berenson@eecs.umich.edu}
}

\begin{document}

\maketitle

\begin{abstract}
Object search -- the problem of finding a target object in a cluttered scene -- is essential to solve for many robotics applications in warehouse and household environments.
However, cluttered environments entail that objects often occlude one another, making it difficult to segment objects and infer their shapes and properties.
Instead of relying on the availability of CAD or other explicit models of scene objects, we augment a manipulation planner for cluttered environments with a state-of-the-art deep neural network for shape completion as well as a volumetric memory system, allowing the robot to reason about what may be contained in occluded areas.
We test the system in a variety of tabletop manipulation scenes composed of household items, highlighting its applicability to realistic domains.
Our results suggest that incorporating both components into a manipulation planning framework significantly reduces the number of actions needed to find a hidden object in dense clutter.
\end{abstract}

\section{Introduction}
\blfootnote{This research was funded in part by Toyota Research Institute (TRI). This article solely reflects the opinions of its authors and not TRI or any other Toyota entity.}
The ability to find and retrieve an object from a cluttered environment, i.e. solving the \textit{object search} problem, is an important requirement for many robotics applications, from warehouse retrieval to household chores. Yet cluttered scenes inherently impose a limitation on visibility: objects occlude one another in close proximity, and often the range of feasible viewpoints is limited. Also, in unstructured environments such as homes, new objects appear frequently and it is very restrictive to require that all objects in the scene have corresponding CAD models and/or labeled images available. In spite of these difficulties, estimating object geometry is important for sequencing actions to find a target object.

Thus this paper focuses on a key topic that has been largely overlooked in the object search domain: inferring occluded geometry. %
The current state-of-the-art in object search (e.g. \cite{jonschkowski2016probabilistic, schwarz2017nimbro, milan2018semantic}), which shows impressive performance, does not reason about occlusion but instead relies on CAD models and/or labeled images to construct a geometric description of the scene. However, when such information is not available, we hypothesize that inferring occluded geometry significantly improves object search performance in dense clutter in terms of the number of actions required to retrieve the object. This paper \textit{does not} focus on the effect of shape-completion on grasping, which has been explored in previous work \cite{varley2017shape}. Rather we focus on the role of shape-completion in action selection, i.e. in determining which object to move and where.

\begin{figure}[t]
    \centering
    \includegraphics[height=1.7in]{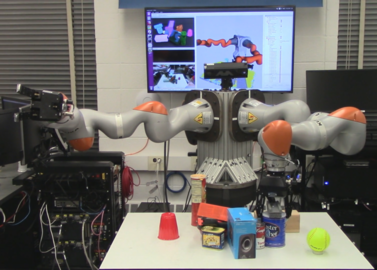}
    \includegraphics[height=1.7in]{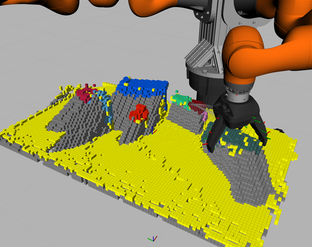}
    \vspace{-1.0\baselineskip}%
    \caption{Left: Our robot sliding an object in a cluttered scene to reveal the target (the yellow ball). Right: The robot's representation of the world, including shape-completed objects and memory of previously seen shapes and free space. Yellow voxels represent the table or unclassified points, gray represents occluded ``shadows'', and other colors represent shape-completed segments. }
    \label{fig:introfig}
    \vspace{-1.0\baselineskip}%
\end{figure}

The main contribution of this paper is the integration of a novel extension to a previous method for shape completion \cite{Yang17} into a manipulation planning framework. Our proposed shape completion method allows us to infer occluded geometry more accurately by using free-space information and we use this method to infer occluded geometry in a cluttered scene. We are not aware of any previous method that integrates shape completion into a framework for object search. We also integrate a memory method to track free space seen earlier in the interaction, as well as previously-observed geometry that has become occluded.

To gauge how much inferring occluded geometry improves manipulation performance, we constructed a manipulation planning system for a bimanual robot (see Figure \ref{fig:introfig}).
Our baseline system consists of a planner that operates on a volumetric segmentation of RGBD images and uses a set of motion primitives to locate and retrieve a target object. We conducted 182 manipulation experiments in eight tabletop cluttered scenes (see Figure \ref{fig:experiment_scenes}) comparing the baseline system to a system augmented with shape completion and/or memory. We first evaluated whether our proposed shape completion method outperformed previous work on a dataset specialized to our application. We then tested the hypothesis that both shape completion and memory independently improved performance in certain scenarios. Finally, we tested the performance of the baseline vs. the augmented system in scenarios with varying amounts of clutter. In densely-cluttered scenes we found that inferring occluded geometry significantly reduced the number of actions necessary to retrieve the target object.

\section{Related Work}

\customsubsubsection{Object Retrieval from Clutter}
Manipulation of movable obstacles in cluttered scenes has been a longstanding goal of robotics research \cite{Stilman2007}, \cite{Dogar2012}.
While these earlier examples assumed full knowledge of the scene in question, leveraging manipulation to discover hidden objects has also been a significant focus.
In \cite{Wong2013}, Wong et al. use spatial visibility constraints and object semantic information to plan manipulation sequences. \cite{Dogar2014} use revealed volume in utility function and generate connected component networks from object occlusions, \cite{Li2016} frame the problem as a POMDP, and employ a similar image processing pipeline to the one presented here. The above methods make important contributions to the literature, however they make simplifying assumptions (e.g. discrete planning space, known object models, or sparse clutter) that are more restrictive than ours.

In recent years, much work about vision and manipulation in clutter has been driven by the scenarios presented by the Amazon Picking Challenge (APC), producing numerous publications on both full frameworks and isolated components.
In APC 2015, \cite{jonschkowski2016probabilistic} developed a segmentation algorithm based on explicit image features (color, edge, missing 3D, distance to shelf, height etc), though the more recent trend has been toward deep-learned perceptual models  (e.g. \cite{schwarz2017nimbro} from APC 2016, combined object detection using a fine-tuned network on Visual Genome and semantic segmentation using a pretrained pixel-level CNN model on ImageNet).
Starting in APC 2017, the challenge required competitors to attempt to manipulate novel items in the scene, producing complex sensing and grasping frameworks such as \cite{milan2018semantic, zeng2018robotic}.
While powerful, these methods do not reason about occluded space.

\customsubsubsection{Volumetric Shape Reconstruction}
Reconstructing a 3D model of a scene is both a major challenge and powerful tool for robotic manipulation.
Until recently, most systems seeking to infer volumetric information about the hidden parts of the scene have relied on CAD model matching \cite{Liu2012Fast, Choi2012} or semantic matching \cite{factored3dTulsiani17, xiang2018posecnn}.
However, progress in reconstructing objects and scenes from single 2.5D views \cite{wu2015shapenets, firman2016structured, song2017semantic, Yang17, fan2017point, Yang18} has enabled manipulation planning on unseen parts of the space \cite{varley2017shape}.
We extend this work in order to reason about occlusion in a cluttered scene, choosing to build on \cite{Yang17} because it obtained good performance on challenging objects and it was clear how to incorporate free-space information into the network.

\section{Problem Statement}

We seek to retrieve a specific object from a cluttered scene using robotic manipulator(s).
Our domain represents household applications, where previously-unseen objects can be present and CAD models or labeled images are not available. 

\customsubsubsection{Sensing}
To sense the environment we assume that the robot is endowed with a single RGB-D sensor. We assume that the objects in the environment are arranged on a flat surface and may be in arbitrary stable pose and contact configurations.
A key difference between our domain and that of much previous work is that we \textit{do not} assume that we possess explicit object representations for the objects on the table: the robot may have observed some of the objects during training of its perception algorithms, but other objects are completely new. Furthermore, the robot has no way to identify objects with which it has been trained. The robot receives an observation of the state of the environment $o_t \in O$ before it acts, where $O$ is an $w \times l$ grid of RGBD values.

The robot also has no explicit representation for the target object, but is endowed with a classifier $L: o_{t} \rightarrow \{0,1\}$ which determines if a given pixel in the RGBD image is likely to be a part of the target object. This is meant to handle queries that may come from a user, such as ``Bring me the yellow ball'', where no explicit model of the object is given.

\customsubsubsection{Acting}
The robot may manipulate the objects in any way it chooses, however, unlike much previous work, we assume we are \textit{not} able to command the robot to remove objects from the scene. We make this restriction to consider realistic scenarios where the robot has a limited work-surface like cabinet interiors or counter tops.

Further, we assume that the robot has only a limited knowledge of contact mechanics and physics. Contacts between the robot model and the environment or between a grasped object and a tabletop object can be computed based on their observed or inferred shapes, but their behavior after contact is difficult to predict because physical properties such as mass, pressure distribution, and friction, are not known.

\customsubsubsection{Problem}
The robot is endowed with a set of possible actions it can apply $\mathcal{A}$ and must choose which actions $a_{1...n} \in \mathcal{A}$ to take to locate and retrieve the target object, with each action $a_i$ having a negative reward and retrieval having a positive reward. This problem can be formulated as a Partially-Observable Markov Decision Process (POMDP) by defining a belief state over the environment and computing a policy of the form $\pi(a_{1:t}, o_{1:t}) = a_{t+1}$ which maximizes the probability of success given any starting state. While a POMDP policy would be desirable, this is clearly intractable as the belief over environments is too high-dimensional for a POMDP solver to handle and we do not have models of the transition and observation uncertainty. Instead, we focus on a greedy approach: we seek to design a $\pi$ that takes the next best action given $a_{1:t}$ and $o_{1:t}$. A key challenge is how to use $a_{1:t}$ and $o_{1:t}$ to infer object geometry in occluded regions, so that this information can be used to inform action selection.

\section{System Framework}

\begin{figure}[t]
    \centering    
    \includegraphics[width=0.75\linewidth]{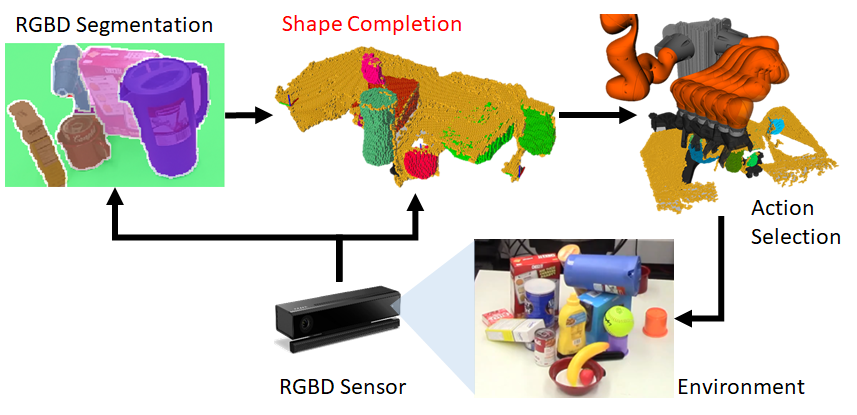}
    \caption{Diagram of data processing pipeline.}
    \label{fig:pipeline}
    \vspace{-2.0\baselineskip}%
\end{figure}

This section describes the components of our system, shown in Figure \ref{fig:pipeline}. We first describe our methods for perception, then action selection.

\subsection{RGB-D Segmentation}

The pipeline begins by processing $o_t$ to produce a segmentation of the observed scene into distinct objects.
Many state-of-the-art semantic segmentation approaches require object class annotations \cite{he2017mask,Lin:2017:RefineNet}, which violates our stipulation that the scene may contain novel objects or classes.
Therefore, we investigated class-agnostic segmentation approaches, adapting SceneCut~\cite{Pham2017} for our scenario.
The method begins with an ultrametric contour map (UCM)~\cite{arbelaez2006boundary}, a hierarchical segmentation tree derived from the RGB-D image of the scene. The UCM is generated by a Convolutional Oriented Boundary  (COB)~\cite{Man+17} network trained on NYUD-v2 dataset~\cite{Silberman:ECCV12}. SceneCut then utilizes a tree cut to minimize an energy function over objectness and geometric fitting.

Each segment of the RGB-D image, given the camera intrinsics, corresponds to a point cloud representing a potential object.
Points belonging to the table surface, robot arms, or outside the table region of interest are rejected, and the surviving point clouds are passed to shape completion. We also try to find the target in $o_t$, as well as extracting occupied and free volumes (detailed below).

\customsubsubsection{Target Object Detection}
Given a collection of image/point cloud segments, we next determine whether any of them is the target in question. We assume a classifier of the form $L:o_t \rightarrow \{0,1\}$ is available, as object recognition is not a focus of this work.
Our implementation uses color matching to classify whether a pixel belongs to the target object. %

\customsubsubsection{Occupied and Free-Space Volumes}
Using the segmentation and full point cloud, we can compute a voxelized representation of the free, occupied, and unknown state of the world.
Occupied and free space computation is provided by feeding the point cloud data to OctoMap \cite{Hornung2013} to generate an octree representation of the scene.
The shape completion results from the following section are fed back into object-specific OctoMaps that are used for end-effector collision computations.

\subsection{Shape Completion}

The main contribution of this paper is the integration of shape completion into our object search system. We first frame the problem of shape completion and then describe our solution. Consider an occupancy map $V \colon \R^3 \to \setof{0, 1}$ carrying 3D points to a binary occupancy value, empty or filled.
Letting $\N_n \defined \setof{0, 1, \ldots, n-1}$ represent the first $n$ natural numbers, we can define a \textit{voxel grid} as a discrete version of $V$, with $V_{n} \colon \N_{n}^3 \to \setof{0, 1}$.

Let $\sparse{V_{n}}$ represent the pointwise representation of $V$: the list of points $\vec{p}$ that are at the center of each voxel and $V(\vec{p})=1$.
We now define a distance between voxel grids based on the \textit{Chamfer distance} as defined by \cite{fan2017point}:
\begin{equation}
\begin{split}
    \lbl{D}{C}(X, Y) \defined
    \frac{1}{\card{X}} \sum_{x \in X} \min_{y \in Y} \norm{x-y}_2^2
    +
    \frac{1}{\card{Y}} \sum_{y \in Y} \min_{x \in X} \norm{x-y}_2^2
\end{split}
\label{eq:chamfer}
\end{equation}
So, for two voxel grids $U_{n}$ and $V_{n}$, we define $D(U_{n}, V_{n}) \defined \lbl{D}{C}(\sparse{U_{n}}, \sparse{V_{n}})$.
Let the true voxel occupancy of an object be $V_{n}^{obj} \subset V_{n}$ and the observed free space be $V_{n}^{free} \subset V_{n}$. Then, given a partial scan of the object $V_{n}^{partial} \subset V_{n}^{obj}$ and $V_{n}^{free}$, shape completion seeks to solve the following problem:

\begin{equation*}
\begin{aligned}
& \underset{V_{n}^{completed}}{\text{argmin}}
& & D(V_{n}^{completed}, V_{n}^{obj}) \\
& \text{subject to}
& & V_{n}^{partial} \subset V_{n}^{completed},\\
& & & V_{n}^{completed} \cap V_{n}^{free} = \emptyset.
\end{aligned}
\end{equation*}

However, at runtime $V_{n}^{obj}$ (the true shape of the object) is unknown. Instead, we apply learning methods which train a deep neural network on multiple views of objects in simulation, where the true shape is used as ground-truth. We then use the learned network to predict a likely $V_{n}^{completed}$ for a given partial scan.

To tackle the learning problem, we begin with a base model of the 3D-RecGAN architecture~\cite{Yang17}, a combination of a generative autoencoder and a Generative Adversarial Network (GAN)\cite{goodfellow2014generative} capable of generating high-resolution 3D shapes that capture key features (such as handles). Compared to previous approaches which generate a 3D shape from RGB or RGB-D information, this architecture does not require object class labels and is able to generalize to unseen objects. This approach performs well, but does not have the ability to include $V_{n}^{free}$, so it may generate voxels in known free space. We thus build on this method by incorporating two main modifications: 1) restructuring the network architecture to include known-free space along the lines of \cite{wu2015shapenets, song2017semantic} and 2) using a dataset that includes occlusions for training.

\begin{figure}[t]
    \centering
    \includegraphics[width=0.95\linewidth]{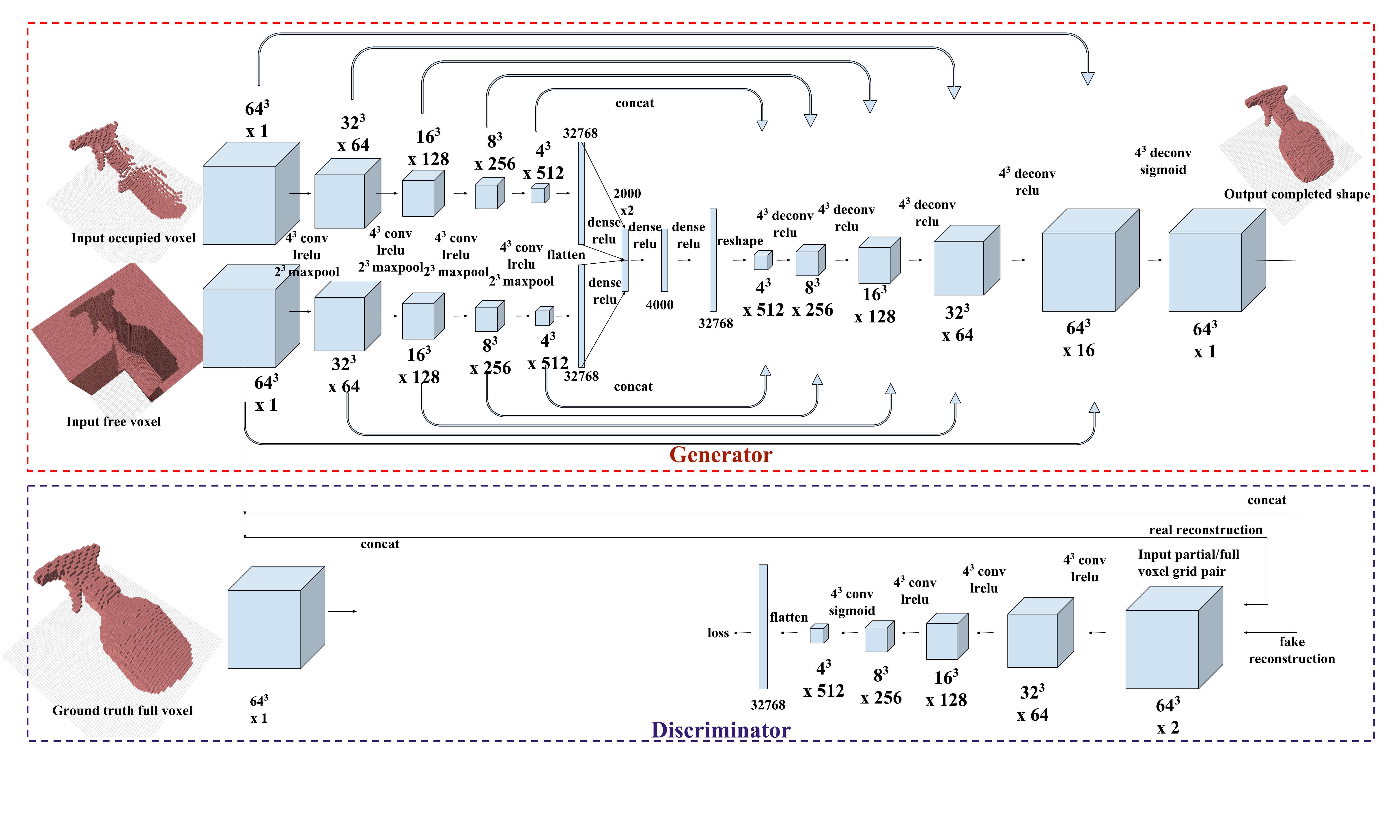}
    \vspace{-2.0\baselineskip}%
    \caption{Proposed network architecture for shape completion.}
    \label{fig:architecture}
    \vspace{-1.0\baselineskip}%
\end{figure}

\customsubsubsection{Architecture}
\label{sec:architecture}
3D-RecGAN consists of two main networks: the generator and the discriminator.
We improve on the original network \cite{Yang17} by augmenting the input space with $V_{n}^{free}$.
Figure \ref{fig:architecture} shows the detailed architecture of our modified generator in 3D-RecGAN.
Both the occupancy voxels and free voxels are encoded using the five 3D convolutional layers used by 3D-RecGAN.
In latent space, the two latent vectors are concatenated together.
The decoder comprises six up-convolutional layers followed by ReLU activations except for the last layer which uses a sigmoid function.
All encoder layers are concatenated to the decoder by skip-connections to preserve local structures.
The discriminator and loss functions are the same as those used in 3D-RecGAN.

\customsubsubsection{Synthesizing a Dataset with Occlusions}
Many large synthetic datasets generated from 3D models exist for the purpose of 3D reconstruction from a single view.
Most existing cluttered datasets are generated on large and complex objects such as furniture in an indoor scene.
Existing datasets for robotic manipulation in cluttered scene are limited.
\cite{varley2017shape} generated a dataset for shape completion for the purpose of robot grasping using objects from the YCB dataset~\cite{calli2015ycb} and Grasping dataset~\cite{kappler2015leveraging}.
However, this dataset only includes a complete single-view occupancy grid with self occlusion only. %
\cite{firman2016structured} introduced a tabletop dataset which consists of a complete RGB-D occluded scene with objects occluding each other, but their TSDF encoding is different from our requirement in the network architecture.

In order to train a network to reconstruct occluded parts from cluttered scenes, we modify and augment the dataset synthesis steps used by~\cite{varley2017shape} so that the objects are not only self-occluded but also occluded by an obstacle. Our dataset contains three kinds of 3D voxel grids for each example: $V_{n}^{partial}$, $V_{n}^{free}$, and $V_{n}^{obj}$ (the ground truth).

12 objects from the YCB dataset and ShapeNet~\cite{shapenet2015} are collected and occupancy grids are generated from the object meshes using binvox~\cite{binvox}.
After that, rotations are uniformly sampled in roll-pitch-yaw space.
Instead of directly generating depth images from different angles of rotations, an obstacle mask is placed in between the camera and the mesh, occluding part of $V_{n}^{obj}$ and thus generating $V_{n}^{partial}$.The voxels between the camera and the $V_{n}^{partial}$ are $V_{n}^{free}$. 
$V_{n}^{partial}$ is then centered in the reconstruction grid in order to remove information about the original object extents so that the input is similar to a real scenario, where the true extent of the object is unclear.
The recentered voxels are then shifted towards the camera to a fixed offset to provide more space for reconstruction.
In the experiment section, we show that training using this occluded data set boosts the performance when reconstructing objects in the presence of occlusion.

\subsection{Volumetric Memory}

Although the dynamics of manipulation are difficult to predict, there are cases, such as when the robot has a stable grasp on a manipulated object, where we wish to inform the next scene of past interactions. With \textit{positive memory}, we compute the pose transformation due to the manipulator motion, then add the object octree at time $t-1$ into the scene octree at time $t$ using the new pose. %
With \textit{negative memory}, we assume that space that was previously free and is now occluded is likely to remain free, unless the shape completion indicates otherwise. %
As both of these assumptions can be violated by unanticipated object interaction (e.g. objects slipping in the grasp or collisions knocking objects behind others), unobserved space is set to decay to the OctoMap occupancy threshold $\lbl{\tau}{occupancy}$ with rate $0 < \alpha < 1$, giving $V_t(\vec{x}) = \alpha V_{t-1}(\vec{x}) + (1-\alpha)\lbl{\tau}{occupancy}$, $\forall \vec{x} \in \operatorname{Unobserved}(V_t)$.

\subsection{Motion Planning}

After segmentation and reconstruction, we are left with a collection of voxel maps approximately representing individual objects which must be rearranged to facilitate target retrieval. We employ a randomized kinodynamic motion planner with heterogeneous action types to find an action to perform in the current scene.
Acting in clutter often restricts the feasibility of traditional pick-and-place actions due to limited reachability around objects. For these reasons, this work follows others, including \cite{Gupta2015}, \cite{Boularias-2015-5904}, \cite{Dogar2012}, in employing action primitives to act in constricted space.

\customsubsubsection{Domain Definitions}
When planning for the motion of rigid bodies in 3D, the natural planning space is the Cartesian product of $\lbl{n}{obj}$ copies of $\mathbb{SE}(3)$, where $\lbl{n}{obj}$ is the number of rigid bodies.
Combined with the robot's joint configuration space $\spc{Q}$, we can form the full configuration space $\spc{C} \defined \mathbb{SE}(3) \times \cdots \times \mathbb{SE}(3) \times \spc{Q}$
representing the state of the robot and all manipulable rigid bodies in the scene.

The action space $\spc{A}$ contains all the possible control actions the system can take.
Each action $a \in \spc{A}$ has an associated parameter space
$\spc{P}_{a}$ and performs the mapping $a \colon \spc{C} \times \spc{P}_{a} \to \spc{C}\label{eq:action_definition}$.
Each action may also be equipped with a steering policy,
$\pi_{a} \colon \spc{C} \times \spc{C} \to \spc{A} \times \spc{P}_{a}\label{eq:steer_policy}$,
which takes an initial and goal configuration and returns the parameterized action to locally advance toward the goal state.
For actions without steering, it is necessary to sample from the parameter space directly.

In the absence of a known terminal state, we instead supply a reward function that determines the most promising action given the current and predicted next state,
$\nu \colon \spc{C} \times \spc{P}_{a} \times \spc{C} \to \mathbb{R}$.
The function $\nu$ rewards actions that are likely to exhibit high information gain and penalizes trajectories with high incidence of collision.
Collisions between the robot and scene objects during action execution are not forbidden in the framework, as the objects are movable, but actions with lower contact are rewarded.

\customsubsubsection{Object Selection}
After segmentation and shape completion, occluded voxels (``shadows'') in the field of view are computed by raycasting from unknown cells back to the camera origin.
The object to move is determined by uniformly sampling from the set of shadow voxels, then casting back to select the object that occluded that volume.
This provides a heuristic for sampling objects that are more likely to be hiding the target object.
In the case where the target object is partially visible, but unreachable due to gripper collisions with its neighbors, the selector has a $\lbl{\tau}{greedy}$ chance of choosing one of those colliding objects (with probability proportional to the number of gripper poses it obstructs and the number of target voxels occluded by the object), and a $1-\lbl{\tau}{greedy}$ chance of proceeding normally. Function SelectObject in Algorithm~\ref{alg:object_selection} highlights this process.

\begin{wrapfigure}{L}{0.5\linewidth}
\begin{minipage}{\linewidth}
\vspace{-3.5\baselineskip}
\begin{algorithm}[H]
    \caption{Motion Generation}
    \label{alg:object_selection}

    \newcommand{\ptilde}{\stackrel{\sim}{\smash{\vec{p}}\rule{0pt}{1.1ex}}}
    
    \begin{algorithmic}
        \Function{ComputeOcclusions}{Octree $T$, Camera $\vec{c}$}
            \State Octree $T_{occl}$
            \ForAll{$\vec{p} \in T$}
                \State $T_{occl}[\vec{p}] \gets ($\Call{RayCast}{$T, \vec{c}, \vec{p}$} $\neq \vec{p})$
            \EndFor
            \State \Return $T_{occl}$
        \EndFunction
        \Function{SelectObject}{Octree $T$, Octree $T_{occl}$, Camera $\vec{c}$, SegmentLookup $S$}
            \State $\ptilde \gets$ \Call{RandomOccupiedNode}{$T_{occl}$}
            \State $\vec{p} \gets$ \Call{RayCast}{$T, \vec{c}, \ptilde$})
            \State \Return $S[\vec{p}]$
        \EndFunction
        \Function{GenMotion}{Parameters $\spc{P}_a$}
            \State PriorityQueue $q$
            \For{$i \in [1 \ldots n_{samples}]$}
                \State $a \gets$ \Call{SampleAction}{$\spc{P}_a$}
                \State $obj \gets$ \Call{SelectObject}{}
                \State $G \gets$ \Call{GetGrasp}{$obj$}
                \State $r \gets \nu(a)$
                \State $\xi \gets$ \Call{GenTrajectory}{$a$, $G$}
                \If{$\xi$}
                    \Call{Push}{$q$, $r$, $\xi$}
                \EndIf
            \EndFor
            \State \Return \Call{Pop}{$q$}
        \EndFunction
    \end{algorithmic}
\end{algorithm}
\vspace{-3.5\baselineskip}
\end{minipage}
\end{wrapfigure}

\customsubsubsection{Action Specifications}
For the tabletop manipulation domain, we have chosen three heterogeneous actions representing a taxonomy based on the controllable subspace of the full state space:
\textsc{Push}, \textsc{Slide}, and \textsc{Pick}.
Each action operates on a single selected object, although in clutter this will likely influence the neighborhood of objects around it.
\textsc{Push}, parameterized by $\spc{P}_{\textsc{Push}}$, represents a 1D palm-push motion with a magnitude and direction in the plane of the table surface.
\textsc{Slide} is implemented by grasping the selected object and dragging in the table plane, for 3 controllable dimensions.
It is parameterized by $\spc{P}_{\textsc{Slide}}$, which contains the $\mathbb{SE}(2)$ transform of the object motion.
Finally, \textsc{Pick} and $\spc{P}_{\textsc{Pick}}$ represent a full grasp of the object, and can move it in $\mathbb{SE}(3)$.

Each task-space motion of an object has a generating policy that produces full joint-space motions of the robot.
This consists of planning a collision free path to the start of the trajectory, then solving coherent inverse kinematics for a Cartesian path of the end-effector.

\begin{figure}[tb]
    \centering
    \subfigure[]{\includegraphics[width=0.5\linewidth]{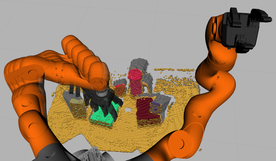}\label{fig:motion_plan}}
    \subfigure[]{\resizebox{!}{11em}{\definecolor{c0000ff}{RGB}{0,0,255}
\definecolor{c808080}{RGB}{128,128,128}

\begin{tikzpicture}[y=0.80pt, x=0.80pt, yscale=-1.000000, xscale=1.000000, inner sep=0pt, outer sep=0pt]
  \path[draw=black,fill=c0000ff,rounded corners=0.0000cm] (132.1429,902.3622)
    rectangle (175.7143,967.3622);
  \path[draw=black,fill=black,line join=miter,line cap=butt,even odd rule,line
    width=0.800pt] (162.8571,696.2908) .. controls (160.0000,718.7908) and
    (191.0714,716.2908) .. (188.2143,731.6479) .. controls (185.3571,747.0051) and
    (213.2143,749.5051) .. (206.7857,728.7908) .. controls (200.3571,708.0765) and
    (205.3571,697.0051) .. (212.8571,692.3622) .. controls (220.3571,687.7194) and
    (218.9286,659.1479) .. (190.3571,669.1479) .. controls (161.7857,679.1479) and
    (162.8571,696.2908) .. (162.8571,696.2908) -- cycle;
  \path[draw=black,dash pattern=on 4.80pt off 0.80pt,line join=miter,line
    cap=butt,miter limit=4.00,even odd rule,line width=0.800pt]
    (254.3440,592.3622) -- (153.9286,902.3622) -- (167.2360,592.3622) --
    (254.3440,592.3622);
  \path[draw=black,fill=c808080,dash pattern=on 0.28pt off 0.28pt,miter
    limit=4.00,line width=0.283pt] (208.1264,731.8553) .. controls
    (208.1066,731.4262) and (207.5350,729.0397) .. (206.8562,726.5520) .. controls
    (202.3978,710.2126) and (204.4132,699.3703) .. (213.0931,692.9986) .. controls
    (215.7041,691.0820) and (216.9912,688.1806) .. (217.1983,683.7449) .. controls
    (217.4649,678.0341) and (215.9939,673.9133) .. (212.5139,670.6221) .. controls
    (210.3268,668.5536) and (208.5114,667.5793) .. (205.5382,666.8783) .. controls
    (199.8714,665.5423) and (192.2541,667.1085) .. (183.0131,671.5099) .. controls
    (174.1854,675.7143) and (168.3973,680.6542) .. (164.9301,686.9429) --
    (163.9090,688.7949) -- (163.9030,687.1429) .. controls (163.8990,686.2343) and
    (164.7716,665.1594) .. (165.8411,640.3099) .. controls (166.9106,615.4604) and
    (167.7857,594.6857) .. (167.7857,594.1439) -- (167.7857,593.1587) --
    (210.6035,593.1587) .. controls (244.7107,593.1587) and (253.3874,593.2486) ..
    (253.2549,593.6006) .. controls (253.1634,593.8437) and (243.5017,623.6350) ..
    (231.7843,659.8035) .. controls (220.0670,695.9720) and (209.9587,727.1554) ..
    (209.3215,729.1000) .. controls (208.6843,731.0445) and (208.1467,732.2844) ..
    (208.1268,731.8553) -- cycle;
\path[draw=black,fill=c0000ff,line join=miter,line cap=butt,even odd rule,line
  width=0.800pt] (153.9286,902.3622) -- (177.8320,861.2872) --
  (129.3216,861.2872) -- cycle;

\end{tikzpicture}} \label{fig:shadow_pose_1}}
    \subfigure[]{\resizebox{!}{11em}{\input{src/shadow_2.tex}} \label{fig:shadow_pose_2}}
    \vspace{-1.0\baselineskip}%
    \caption{Motion planning scene and motion rewards. \ref{fig:motion_plan} Unclassified occupied space is represented with gold voxels, shape-completed segments with random colors, and shadow voxels with gray. A \textsc{Slide} motion primitive is displayed with the robot motion trail. The right hand side shows dis-occluded voxels after an object motion. \ref{fig:shadow_pose_1} shows a shape with its shadow given a camera pose. \ref{fig:shadow_pose_2} shows the scene after an object motion, with the newly visible regions highlighted in green. The volume of these regions determines the reward.}
    \label{fig:motion_scene}
\end{figure}

\customsubsubsection{Feasibility Function}
The feasibility function $\Phi_a$ is composed of two components.
First, the entire generated trajectory must be kinematically feasible.
Second, the initial pose of the hand must be free from collisions. 
Collisions between the remainder of the robot trajectory and the scene objects are permitted, and are handled by the reward function.

\customsubsubsection{Reward Function}
Given limited knowledge of the scene's dynamics and state, the reward function $\nu$ plays a dominant role in enabling progress toward locating the target object.
The reward used here is a linear combination of a number of heuristic value functions $\nu_i$: $\nu = \begin{bmatrix}\nu_1 & \nu_2 & \nu_3 & \nu_4\end{bmatrix} \vec{w} \label{eq:composite_reward}$.

For our constituent reward functions, we use the following elements:
\begin{description}[labelwidth=2cm]
    \item[Information] $\nu_1$: The number of previously occluded voxels that should be revealed by this motion.
    \item[Dispersion] $\nu_2$: The standard deviation of the centroids of detected objects.
    \item[Direction] $\nu_3$: The motion of the object toward or away from the center of mass of the scene.
    \item[Collision] $\nu_4$: The number of collisions between the robot trajectory and non-manipulated objects in the current motion.
\end{description}
Heuristic $\nu_1$ is the dominant value, representing how much of the scene is revealed as shown in Figure~\ref{fig:motion_scene}.
Heuristics $\nu_{2-4}$ primarily act to encourage spreading objects apart, which assists in object search and grasp generation.
No penalty is assessed for disturbing or toppling other objects, other than the collision metric.
The weight vector $\vec{w}$ will in general depend highly on the resolution of the voxel map.

The Information heuristic requires some additional explanation.
When moving a partially-visible object with shadow, the hidden voxels could ``belong'' either to the object in motion or to the remainder of the scene.
If the object motion is represented by a rigid transform $T^{\vec{s}}_{\vec{s}^\prime}$ and a shadowed voxel coordinate by $\lbl{\vec{p}}{occl}$, then we need to check whether either $\lbl{\vec{p}}{occl}$ or $\lbl{\vec{p}^\prime}{occl}=T^{\vec{s}}_{\vec{s}^\prime}\lbl{\vec{p}}{occl}$ are visible given the new object position.
Figures~\ref{fig:shadow_pose_1} and \ref{fig:shadow_pose_2} show this process.

\customsubsubsection{Grasp Generation}
For both the cases with and without shape completion, grasps were planned on a convex prismatic approximation of the voxel grid, generated by extruding the 2D convex hull of the object's downprojected (X-Y) occupancy map between its minimum and maximum extents in the table z-coordinate.

\section{Experimental Results}

\begin{figure*}[t]
    \centering
    \subfigure[A]{\includegraphics[width=0.22\linewidth]{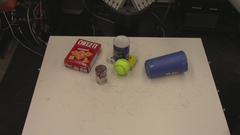} \label{fig:scene_A}}
    \subfigure[B]{\includegraphics[width=0.22\linewidth]{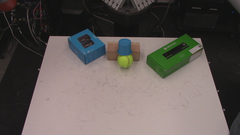} \label{fig:scene_B}}
    \subfigure[C1]{\includegraphics[width=0.22\linewidth]{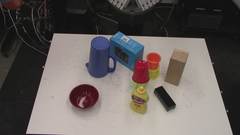} \label{fig:scene_C1}}
    \subfigure[C2]{\includegraphics[width=0.22\linewidth]{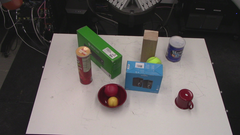} \label{fig:scene_C2}}
    \subfigure[C3]{\includegraphics[width=0.22\linewidth]{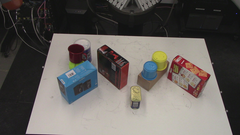} \label{fig:scene_C3}}
    \subfigure[D1]{\includegraphics[width=0.22\linewidth]{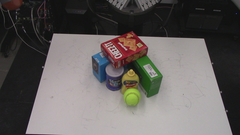} \label{fig:scene_D1}}
    \subfigure[D2]{\includegraphics[width=0.22\linewidth]{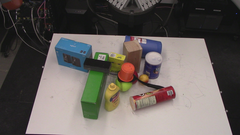} \label{fig:scene_D2}}
    \subfigure[D3]{\includegraphics[width=0.22\linewidth]{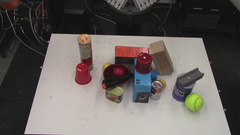} \label{fig:scene_D3}}
    \vspace{-1.0\baselineskip}%
    \caption{Experimental configurations. In each scene, the target object is the yellow/green softball. Occluding objects are roughly 75\% YCB, with the remainder previously unseen by the shape completer. The robot is located at the top of the image where it cannot see the softball.}
    \label{fig:experiment_scenes}
    \vspace{-1.0\baselineskip}%
\end{figure*}

\customsubsubsection{Experimental Equipment}
To evaluate the system described, we employed a custom bimanual robot equipped with two KUKA LBR iiwa 7 R800 arms, two Robotiq 3-Finger Adaptive Grippers, and a Microsoft Kinect 2 for vision.
External localization of the robot, camera, and table was provided by eight Vicon Bonita 10 motion capture cameras (no motion capture was used for the manipulated objects).
Primary scene processing and motion planning was performed on a PC with an Intel 4.7GHz i7-8700K CPU and NVIDIA GTX 1080Ti GPU.
Shape completion was also performed on a 1080Ti, and segmentation was performed on an NVIDIA Tesla V100-SXM2.
The system, spanning six PCs including hardware-facing machines, used ROS for interprocess communication, sensor data acquisition, and trajectory transmission.

\customsubsubsection{Experimental Parameters}
Throughout the preceding sections, several threshold and weight parameters were employed. For this experiment configuration, we used values of $\lbl{\tau}{target} = 0.5$, $\lbl{\tau}{greedy} = 0.9$, $\lbl{\tau}{occupancy} = 0.5$, and from Equation~\ref{eq:composite_reward} $\vec{w} = \begin{bmatrix} \tfrac{1}{2000} & 1 & 3 & -5 \end{bmatrix}^\transpose$.

\subsection{Shape Completion}

\customsubsubsection{Experiment setup}
To benchmark the shape completion modifications, we generated voxel grids for 16 objects in a variety of previously unseen orientations, and where four of the objects were previously unseen by the network.
One quadrant of the view was then occluded to mimic the conditions found in realistic scenes.
864 data points are generated for each object and randomly split into training and testing sets with the ratio of 4:1.
The testing dataset also include 3 new objects which were not in the training set.
The resulting reconstructions were then compared using the chamfer distance metric from Equation~\ref{eq:chamfer}.
These error statistics are collected for all 8,448 data points in Figure~\ref{fig:shape_completion_boxplot}. We then evaluated two hypotheses about our shape completion method:

\begin{figure}[t]
    \centering
    \resizebox{\linewidth}{!}{\input{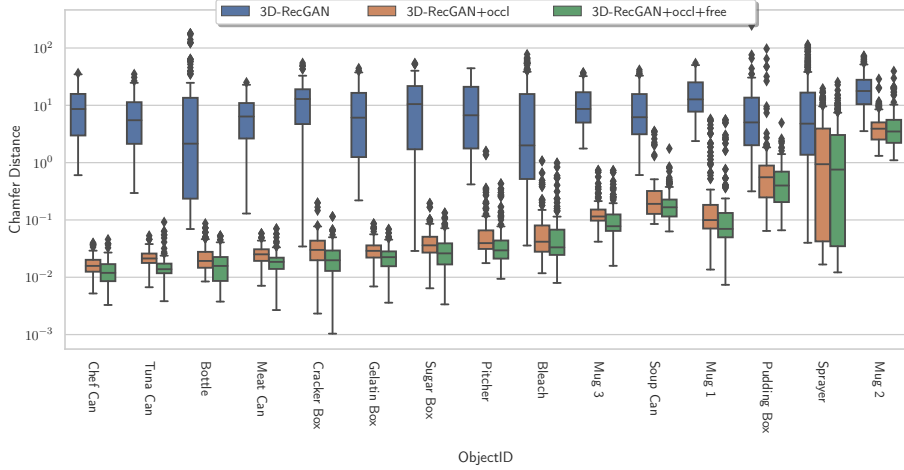}}
    \vspace{-1.0\baselineskip}%
    \caption{Chamfer distances of various shape completion methods.}
    \label{fig:shape_completion_boxplot}
    \vspace{-1.0\baselineskip}%
\end{figure}

\begin{figure}[t]
    \centering
    \subfigure
    {
        \includegraphics[width=0.19\linewidth]{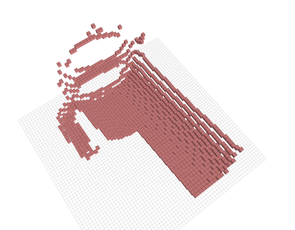}
        \includegraphics[width=0.19\linewidth]{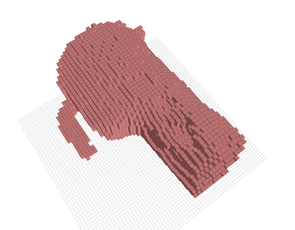}
        \includegraphics[width=0.19\linewidth]{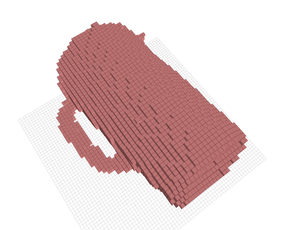}
        \includegraphics[width=0.19\linewidth]{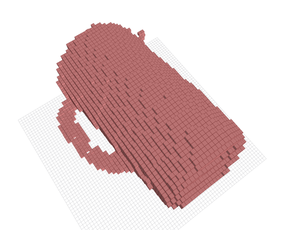}
        \includegraphics[width=0.19\linewidth]{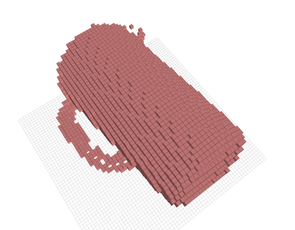}
    }\\ \vspace{-1em}%
    \subfigure
    {
        \includegraphics[width=0.19\linewidth]{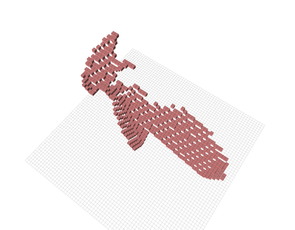}
        \includegraphics[width=0.19\linewidth]{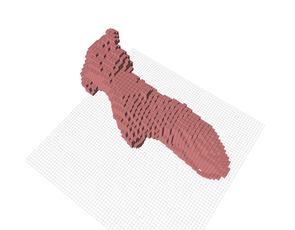}
        \includegraphics[width=0.19\linewidth]{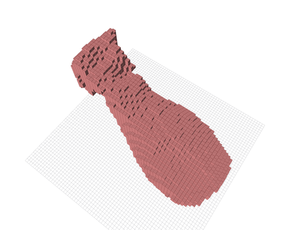}
        \includegraphics[width=0.19\linewidth]{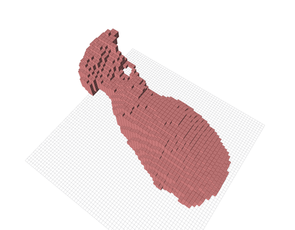}
        \includegraphics[width=0.19\linewidth]{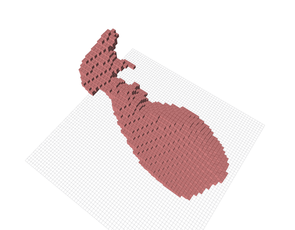}
    }\\ \vspace{-1em}%
    \subfigure
    {
        \includegraphics[                    width=0.19\linewidth]{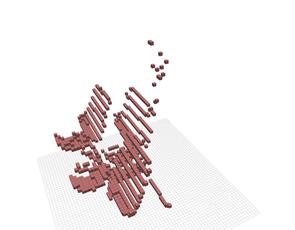}
        \includegraphics[width=0.19\linewidth]{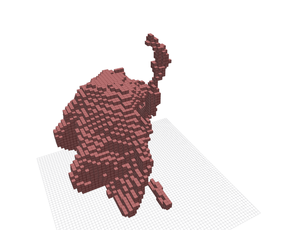}
        \includegraphics[width=0.19\linewidth]{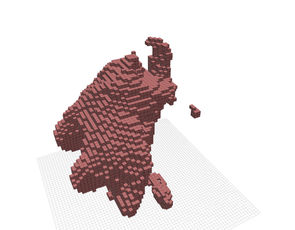}
        \includegraphics[width=0.19\linewidth]{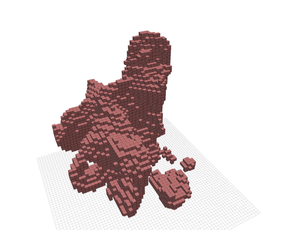}
        \includegraphics[width=0.19\linewidth]{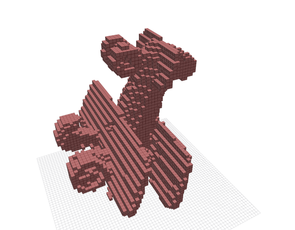}
    }\\
    \subfigure
    {
        (a)\hspace{0.16\linewidth}(b)\hspace{0.16\linewidth}(c)\hspace{0.16\linewidth}(d)\hspace{0.16\linewidth}(e)
    }
    \vspace{-0.2in}
    \caption{Visualization of results of shape completion on several objects. (a) Input voxels; (b) 3D-RecGAN trained on unoccluded dataset; (c) 3D-RecGAN trained on occluded dataset; (d) 3D-RecGAN trained on occluded dataset with free space as augmented input. (e) Ground truth. The pitcher is in the training set. The sprayer and toy airplane are not in the training set.}
    \label{fig:visual_recon_on_diff_dataset}
\end{figure}

\begin{hyp} 
Training shape completion with a dataset of occluded objects significantly improves performance.
\end{hyp}
As seen in Figure~\ref{fig:shape_completion_boxplot}, augmenting the training data with occlusions provides a dramatic improvement in performance (note the log scale on the y-axis).
Comparing the 3D-RecGAN model with the 3D-RecGAN + occlusions model yields a t-statistic of 36.752 and a p-value of $\approx 0$.
Examples are shown in Figure~\ref{fig:visual_recon_on_diff_dataset}.
This hypothesis is strongly supported by the results.

\begin{hyp} 
Including known free voxel information in shape completion significantly improves performance.
\end{hyp}
On top of the additions to the training dataset, Section \ref{sec:architecture} described a modification to the original 3D-RecGAN architecture to account for known free space.
Figure~\ref{fig:shape_completion_boxplot} shows these results as well, showing a modest improvement from the non-freespace network.
Computing the t-statistics for the freespace and non-freespace leads to a t-value of 1.87543, and a p-value of 0.06079.

\subsection{Tests in manually-designed scenes}

In order to show the capabilities of shape completion and memory, we test each of these components in manually-designed scenarios where that component is beneficial. We then compare the results to using the baseline (the framework without either of these components). These test scenarios are shown in Figures~\ref{fig:scene_A}~and~\ref{fig:scene_B}.

In all scenes, the target object is the yellow ball, and the scene is considered to be successfully solved when the robot picks the ball from the scene using either hand.
An attempt is marked as a failure if the target object is ejected from the workspace during the course of the attempt, or if the number of actions taken is more than three times the number of objects in the scene.

\begin{figure}[t]
    \centering
    \resizebox{0.99\linewidth}{!}{\input{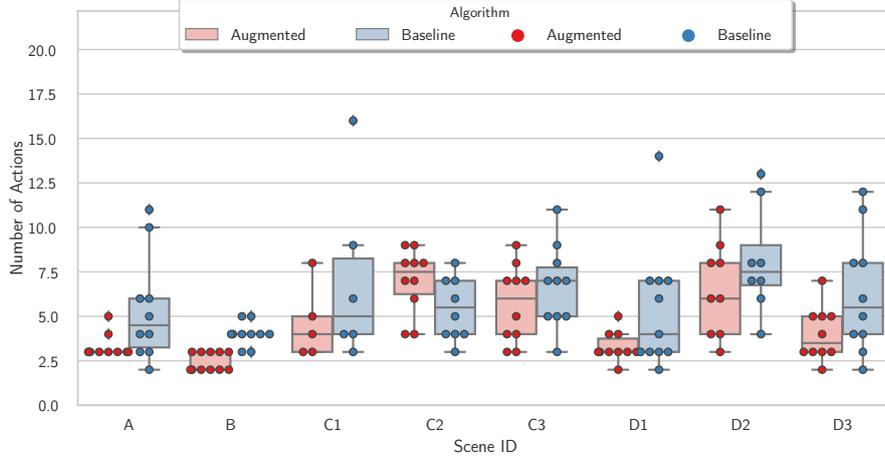}}
    \vspace{-1.0\baselineskip}%
    \caption{Number of actions executed to retrieve the target object from each scene.}
    \label{fig:experiment_results}
    \vspace{-1.0\baselineskip}%
\end{figure}

\begin{hyp}
Memory significantly reduces the number of actions necessary to retrieve a target object when other objects are likely to be investigated first.
\end{hyp}
Scene A was designed to explore the benefits of volumetric memory in locating a hidden object in the scene.
Here the pitcher casts a much larger shadow than the coffee can behind which the target is hiding, but after one or two moves the system should realize that the target is not there, and should prioritize other objects.
Figure~\ref{fig:experiment_results} and Table~\ref{tab:experiment_results} show strong improvement, with a t-statistic of 5.8 and a p-value of 0.00002.
This furnishes a compelling argument that memory greatly improves the system performance.

\begin{hyp}
Shape completion significantly reduces the number of actions necessary to retrieve a target object when visible objects cause large occlusions.%
\end{hyp}
Scene B was constructed to demonstrate the capability of shape completion to rule out some of the occluded area for exploration because it is  a part of visible objects. 
In the scenario, the target is hidden behind a small cylinder, while there are two large boxes as distractions. The boxes have large shadows, so the baseline would be biased to move those to find the hidden object.
However, with accurate shape completion, the system should realize that most of the volume shadowed by the boxes is likely part of the boxes themselves, and thus should move the cylinder.
Figure~\ref{fig:experiment_results} shows that about half of the time the augmented system decides to move the cylinder first, retrieving the target in the optimal two moves. In the other cases, the augmented system selects one of the side boxes first, or needs to clear an obstacle in contact with the target before grasping.
Table~\ref{tab:experiment_results} shows improvement over the baseline, with a t-statistic of 2.1 and a p-value of 0.056.

\subsection{Tests in arbitrary cluttered scenes}

To assess the performance of the framework as a whole (including both memory and shape completion), we tested the full framework vs. the baseline (without shape completion or memory) on arbitrary sparsely-cluttered scenes and densely-cluttered scenes. We generated a collection of arbitrary sparse and dense clutter scenes, shown in Figure~\ref{fig:experiment_scenes} as C1-C3 and D1-D3.
For our purposes, ``dense'' clutter is defined to be where most or all objects in the scene are in contact with one another.

\begin{hyp} Our full framework significantly reduces the number of actions necessary to retrieve a target object in sparsely-cluttered scenarios.
\end{hyp}
Scenes C1-3 were constructed to resemble typical household clutter.
Figure~\ref{fig:experiment_results} and Table~\ref{tab:experiment_results} show strong improvement on C1 and C3, but C2 shows a minor regression.
Thus, the hypothesis is only weakly supported, with a t-statistic of 1.5 and a p-value of 0.14.

\begin{hyp} Our full framework significantly reduces the number of actions necessary to retrieve a target object in densely-cluttered scenarios.
\end{hyp}
Scenes D1-3 were constructed to resemble typical household clutter that is more densely distributed than C.
In all of these cases, the augmented system showed strong improvement, with a t-statistic of 2.6 and a p-value of 0.012, showing good support for the hypothesis.

\begin{wraptable}{r}{0.7\linewidth}
    \centering
    \vspace{-2.0\baselineskip}%
    \caption{Performance Statistics of Augmented vs. Baseline}
    \begin{tabular}{lrrrrrr}
\toprule
{} & \multicolumn{4}{c}{Moves} & \multicolumn{2}{c}{Success Ratio} \\
{} & \multicolumn{2}{c}{Augmented} & \multicolumn{2}{c}{Baseline} \\
{} &            Mean &            StdErr &           Mean &           StdErr &  Augmented &  Baseline \\
\midrule
A  &  \textbf{3.300} &             0.213 &          5.400 &            0.945 &              1.000 &             1.000 \\
B  &  \textbf{2.455} &             0.157 &          4.000 &            0.211 &              1.000 &             1.000 \\
C1 &  \textbf{4.600} &             0.927 &          7.000 &            2.000 &              0.833 &    \textbf{1.000} \\
C2 &           7.000 &             0.577 &          6.900 &            1.650 &     \textbf{1.000} &             0.769 \\
C3 &  \textbf{5.700} &             0.684 &          6.700 &            0.731 &     \textbf{1.000} &             0.909 \\
D1 &  \textbf{3.300} &             0.260 &          5.364 &            1.038 &     \textbf{1.000} &             0.917 \\
D2 &  \textbf{6.556} &             0.884 &          8.125 &            1.060 &     \textbf{0.900} &             0.800 \\
D3 &  \textbf{4.000} &             0.471 &          6.300 &            1.065 &              1.000 &             1.000 \\
\bottomrule
\end{tabular}
    \label{tab:experiment_results}
    \vspace{-2.0\baselineskip}%
\end{wraptable}

\subsection{Computation Time}

Our framework is a proof-of-concept and has not been optimized for fast computation or execution. However, to gauge the practicality of our method, we collected statistics on average computation time used for each component: Preprocessing: 4.34s; Segmentation: 7.30s; Shape Completion: 1.67s; Memory $\approx 0$; Action Selection: 7.32s, and Execution: 34.98s. These results show the the benefits of shape completion and memory come at a low computational cost as compared to the rest of the framework.

\section{Conclusions and Future Work}

This paper has presented a method for the volumetric completion of partially observed scenes and demonstrated that such a method, when integrated with a manipulation planner, can significantly reduce the number of actions required to retrieve a hidden object from dense clutter.
In addition, we have shown that our extension of previous work on shape completion to consider partially-occluded views and known free-space volumes can boost the performance of shape completion in cluttered environments. Future work focuses on enhancing the segmentation component by inferring better segmentations from video of interactions.

\printbibliography

@inproceedings{Silberman:ECCV12,
  author    = {Nathan Silberman, Derek Hoiem, Pushmeet Kohli and Fergus, Rob},
  booktitle = {{ECCV}},
  date      = {2012},
  title     = {Indoor Segmentation and Support Inference from RGBD Images},
}

@article{Man+17,
  author       = {Maninis, K.K. and Pont-Tuset, J. and Arbel\'{a}ez, P. and Gool, L. Van},
  date         = {2017},
  journaltitle = {{TPAMI}},
  title        = {Convolutional Oriented Boundaries: From Image Segmentation to High-Level Tasks},
}

@inproceedings{arbelaez2006boundary,
  author       = {Arbelaez, Pablo},
  organization = {IEEE},
  booktitle    = {{CVPR Workshop}},
  date         = {2006},
  pages        = {182--182},
  title        = {Boundary extraction in natural images using ultrametric contour maps},
}

@inproceedings{song2017semantic,
  title={Semantic scene completion from a single depth image},
  author={Song, Shuran and Yu, Fisher and Zeng, Andy and Chang, Angel X and Savva, Manolis and Funkhouser, Thomas},
  booktitle={Proceedings of the IEEE Conference on Computer Vision and Pattern Recognition},
  pages={1746--1754},
  year={2017}
}

@inproceedings{Pham2017,
  author    = {{Pham}, T. and {Do}, T.-T. and {S{ü}nderhauf}, N. and {Reid}, I.},
  booktitle = {{ICRA}},
  date      = {2018},
  title     = {{SceneCut: Joint Geometric and Object Segmentation for Indoor Scenes}},
}

@inproceedings{Lin:2017:RefineNet,
  author     = {Lin, G. and Milan, A. and Shen, C. and Reid, I.},
  booktitle  = {{CVPR}},
  date       = {2017-07},
  shorttitle = {RefineNet: Multi-Path Refinement Networks},
  title      = {Refine{N}et: {M}ulti-Path Refinement Networks for High-Resolution Semantic Segmentation},
}

@inproceedings{he2017mask,
  author       = {He, Kaiming and Gkioxari, Georgia and Doll{á}r, Piotr and Girshick, Ross},
  organization = {IEEE},
  booktitle    = {{ICCV}},
  date         = {2017},
  pages        = {2980--2988},
  title        = {Mask r-cnn},
}

@article{Stilman2007,
  author       = {Stilman, Mike and Schamburek, Jan Ullrich and Kuffner, James and Asfour, Tamim},
  date         = {2007},
%   doi          = {10.1109/ROBOT.2007.363986},
%   issn         = {10504729},
  journaltitle = {{ICRA}},
  pages        = {3327--3332},
  title        = {{Manipulation planning among movable obstacles}},
}

@article{Dogar2012,
  author       = {Dogar, Mehmet R. and Srinivasa, Siddhartha S.},
  date         = {2012},
%   doi          = {10.1007/s10514-012-9306-z},
%   issn         = {09295593},
  journaltitle = {Autonomous Robots},
  number       = {3},
  pages        = {217--236},
  title        = {{A planning framework for non-prehensile manipulation under clutter and uncertainty}},
  volume       = {33},
}

@article{Wong2013,
  author       = {Wong, Lawson L.S. and Kaelbling, Leslie Pack and Lozano-Perez, Tomas},
  date         = {2013},
%   doi          = {10.1109/ICRA.2013.6630966},
%   issn         = {10504729},
  journaltitle = {{ICRA}},
  pages        = {2814--2819},
  title        = {{Manipulation-based active search for occluded objects}},
}

@article{Dogar2014,
  author       = {Dogar, Mehmet R. and Koval, Michael C. and Tallavajhula, Abhijeet and Srinivasa, Siddhartha S.},
  date         = {2014},
%   doi          = {10.1007/s10514-013-9372-x},
%   issn         = {09295593},
  journaltitle = {Autonomous Robots},
  number       = {1-2},
  pages        = {153--167},
  title        = {{Object search by manipulation}},
  volume       = {36},
}

@article{Gupta2015,
  author       = {Gupta, Megha and M{ü}ller, J{ö}rg and Sukhatme, Gaurav S.},
  date         = {2015},
%   doi          = {10.1109/TASE.2014.2361346},
%   issn         = {15455955},
  journaltitle = {IEEE Transactions on Automation Science and Engineering},
  number       = {2},
  pages        = {608--614},
  title        = {{Using Manipulation Primitives for Object Sorting in Cluttered Environments}},
  volume       = {12},
}

@article{Li2016,
  author       = {Li, Jue Kun and Hsu, David and Lee, Wee Sun},
  date         = {2016},
%   doi          = {10.1109/IROS.2016.7759839},
%   issn         = {21530866},
  journaltitle = {{IROS}},
  pages        = {5701--5707},
  title        = {{Act to see and see to act: POMDP planning for objects search in clutter}},
  volume       = {2016-November},
}

@article{Liu2012Fast,
  author       = {Liu, Ming Yu and Tuzel, Oncel and Veeraraghavan, Ashok and Taguchi, Yuichi and Marks, Tim K. and Chellappa, Rama},
  date         = {2012},
%   doi          = {10.1177/0278364911436018},
%   issn         = {02783649},
  journaltitle = {{IJRR}},
  number       = {8},
  pages        = {951--973},
  title        = {{Fast object localization and pose estimation in heavy clutter for robotic bin picking}},
  volume       = {31},
}

@article{Choi2012,
  author       = {Choi, Changhyun and Christensen, Henrik Iskov},
  date         = {2012},
  journaltitle = {{IROS}},
  title        = {{3D pose estimation of daily objects using an RGB-D camera}},
}

@inproceedings{Boularias-2015-5904,
  author    = {Boularias, Abdeslam and Bagnell, J. Andrew and Stentz, Anthony},
  booktitle = {{AAAI}},
  date      = {2015-01},
  title     = {Learning to Manipulate Unknown Objects in Clutter by Reinforcement},
}

@inproceedings{jonschkowski2016probabilistic,
  author       = {Jonschkowski, Rico and Eppner, Clemens and H{ö}fer, Sebastian and Mart{í}n-Mart{í}n, Roberto and Brock, Oliver},
  organization = {IEEE},
  booktitle    = {{IROS}},
  date         = {2016},
  pages        = {1--7},
  title        = {Probabilistic multi-class segmentation for the amazon picking challenge},
}

@inproceedings{schwarz2017nimbro,
  author       = {Schwarz, Max and Milan, Anton and Lenz, Christian and Munoz, Aura and Periyasamy, Arul Selvam and Schreiber, Michael and Sch{ü}ller, Sebastian and Behnke, Sven},
  organization = {IEEE},
  booktitle    = {{ICRA}},
  date         = {2017},
  pages        = {3032--3039},
  title        = {NimbRo Picking: Versatile part handling for warehouse automation},
}

@inproceedings{milan2018semantic,
  author       = {Milan, Anton and Pham, Trung and Vijay, K and Morrison, Douglas and Tow, Adam W and Liu, L and Erskine, J and Grinover, R and Gurman, A and Hunn, T and others},
  organization = {IEEE},
  booktitle    = {{ICRA}},
  date         = {2018},
  pages        = {1908--1915},
  title        = {Semantic segmentation from limited training data},
}

@inproceedings{zeng2018robotic,
  author    = {Zeng, Andy and Song, Shuran and Yu, Kuan-Ting and Donlon, Elliott and Hogan, Francois Robert and Bauza, Maria and Ma, Daolin and Taylor, Orion and Liu, Melody and Romo, Eudald and Fazeli, Nima and Alet, Ferran and Dafle, Nikhil Chavan and Holladay, Rachel and Morona, Isabella and Nair, Prem Qu and Green, Druck and Taylor, Ian and Liu, Weber and Funkhouser, Thomas and Rodriguez, Alberto},
  booktitle = {{ICRA}},
  date      = {2018},
  title     = {Robotic Pick-and-Place of Novel Objects in Clutter with Multi-Affordance Grasping and Cross-Domain Image Matching},
}

@inproceedings{goodfellow2014generative,
  author    = {Goodfellow, Ian and Pouget-Abadie, Jean and Mirza, Mehdi and Xu, Bing and Warde-Farley, David and Ozair, Sherjil and Courville, Aaron and Bengio, Yoshua},
  booktitle = {Advances in neural information processing systems},
  date      = {2014},
  pages     = {2672--2680},
  title     = {Generative adversarial nets},
}

@inproceedings{wu2015shapenets,
  author    = {Zhirong Wu and Shuran Song and Aditya Khosla and Fisher Yu and Linguang Zhang and Xiaoou Tang and Jianxiong Xiao},
  booktitle = {{CVPR}},
  date      = {2015},
  pages     = {1912--1920},
  title     = {{3D ShapeNets: A Deep Representation for Volumetric Shapes}},
}

@inproceedings{factored3dTulsiani17,
  author    = {Tulsiani, Shubham and Gupta, Saurabh and Fouhey, David and Efros, Alexei A. and Malik, Jitendra},
  booktitle = {{CVPR}},
  date      = {2018},
  title     = {Factoring Shape, Pose, and Layout from the 2D Image of a 3D Scene},
}

@inproceedings{xiang2018posecnn,
  author    = {Xiang, Yu and Schmidt, Tanner and Narayanan, Venkatraman and Fox, Dieter},
  date      = {2018},
  booktitle = {Robotics: Science and Systems (RSS)},
  title     = {PoseCNN: A Convolutional Neural Network for 6D Object Pose Estimation in Cluttered Scenes},
}

@inproceedings{fan2017point,
  author    = {Fan, Haoqiang and Su, Hao and Guibas, Leonidas J},
  booktitle = {{CVPR}},
  date      = {2017},
  number    = {4},
  pages     = {6},
  title     = {A Point Set Generation Network for 3D Object Reconstruction from a Single Image.},
  volume    = {2},
}

@inproceedings{firman2016structured,
  author    = {Firman, Michael and Mac Aodha, Oisin and Julier, Simon and Brostow, Gabriel J},
  booktitle = {{CVPR}},
  date      = {2016},
  pages     = {5431--5440},
  title     = {Structured prediction of unobserved voxels from a single depth image},
}

@inproceedings{Yang18,
  author    = {Yang, Bo and Rosa, Stefano and Markham, Andrew and Trigoni, Niki and Wen, Hongkai},
  booktitle = {{TPAMI}},
  date      = {2018},
  title     = {Dense 3D Object Reconstruction from a Single Depth View},
}

@inproceedings{Yang17,
  author    = {Yang, Bo and Wen, Hongkai and Wang, Sen and Clark, Ronald and Markham, Andrew and Trigoni, Niki},
  booktitle = {{ICCV Workshop}},
  date      = {2017},
  title     = {3D Object Reconstruction from a Single Depth View with Adversarial Learning},
}

@inproceedings{varley2017shape,
  author       = {Varley, Jacob and DeChant, Chad and Richardson, Adam and Ruales, Joaqu{ı́}n and Allen, Peter},
  organization = {IEEE},
  booktitle    = {{IROS}},
  date         = {2017},
  pages        = {2442--2447},
  title        = {Shape completion enabled robotic grasping},
}

@inproceedings{calli2015ycb,
  author       = {Calli, Berk and Singh, Arjun and Walsman, Aaron and Srinivasa, Siddhartha and Abbeel, Pieter and Dollar, Aaron M},
  organization = {IEEE},
  booktitle    = {{ICAR}},
  date         = {2015},
  pages        = {510--517},
  title        = {The YCB object and model set: Towards common benchmarks for manipulation research},
}

@inproceedings{kappler2015leveraging,
  author       = {Kappler, Daniel and Bohg, Jeannette and Schaal, Stefan},
  organization = {IEEE},
  booktitle    = {{ICRA}},
  date         = {2015},
  pages        = {4304--4311},
  title        = {Leveraging big data for grasp planning},
}

@report{shapenet2015,
  author      = {Chang, Angel X. and Funkhouser, Thomas and Guibas, Leonidas and Hanrahan, Pat and Huang, Qixing and Li, Zimo and Savarese, Silvio and Savva, Manolis and Song, Shuran and Su, Hao and Xiao, Jianxiong and Yi, Li and Yu, Fisher},
%   institution = {Stanford University --- Princeton University --- Toyota Technological Institute at Chicago},
  date        = {2015},
  number      = {arXiv:1512.03012},
  title       = {{ShapeNet: An Information-Rich 3D Model Repository}},
%   type        = {techreport},
}

@article{Hornung2013,
  author       = {Hornung, Armin and Wurm, Kai M. and Bennewitz, Maren and Stachniss, Cyrill and Burgard, Wolfram},
  date         = {2013-02},
%   doi          = {10.1007/s10514-012-9321-0},
%   issn         = {0929-5593},
  journaltitle = {Autonomous Robots},
  number       = {3},
  pages        = {189--206},
  title        = {{OctoMap: an efficient probabilistic 3D mapping framework based on octrees}},
  volume       = {34},
}

@misc{binvox,
  author       = {Min, Patrick},
  date         = {2004},
  howpublished = {\url{http://www.patrickmin.com/binvox}},
  note         = {Accessed: 2018-05-01},
  title        = {binvox},
}

\end{document}